\documentclass[10pt,letterpaper]{article}

\usepackage{cogsci}

\cogscifinalcopy 

\usepackage{pslatex}
\usepackage{apacite}
\usepackage{float} 
\usepackage{graphicx}

\usepackage{color}

\title{Robustness of Object Recognition under Extreme Occlusion\\in Humans and Computational Models}
 
\author{{\large \bf Hongru Zhu$^{1}$ \ \ \ \ Peng Tang$^{2}$ \ \ \ \ Jeongho Park$^{3}$ \ \ \ \ Soojin Park$^{4}$ \ \ \ \ Alan Yuille$^{1, 2}$} \\
$^{1}$Department of Cognitive Science, The Johns Hopkins University, Baltimore, MD 21218 USA\\
$^{2}$Department of Computer Science, The Johns Hopkins University, Baltimore, MD 21218 USA\\
$^{3}$Department of Psychology, Harvard University, Cambridge, MA 02138 USA\\
$^{4}$Department of Psychology, Yonsei University, Seoul, Republic of Korea\\
{\tt\small hzhu38@jhu.edu \ \ \ \ \{tangpeng723,p9j8h7,sjpark31,alan.l.yuille\}@gmail.com}
}
 
\begin{document}

\maketitle

\begin{abstract}
Most objects in the visual world are partially occluded, but humans can recognize them without difficulty.
However, it remains unknown whether object recognition models like convolutional neural networks (CNNs) can handle real-world occlusion.
It is also a question whether efforts to make these models robust to constant mask occlusion are effective for real-world occlusion. We test both humans and the above-mentioned computational models in a challenging task of object recognition under extreme occlusion, where target objects are heavily occluded by irrelevant real objects in real backgrounds. Our results show that human vision is very robust to extreme occlusion while CNNs are not, even with modifications to handle constant mask occlusion. This implies that the ability to handle constant mask occlusion does not entail robustness to real-world occlusion. As a comparison, we propose another computational model that utilizes object parts/subparts in a compositional manner to build robustness to occlusion. This performs significantly better than CNN-based models on our task with error patterns similar to humans. These findings suggest that testing under extreme occlusion can better reveal the robustness of visual recognition, and that the principle of composition can encourage such robustness.


\textbf{Keywords:} 
visual recognition; occlusion; computational model; neural network; psychophysics
\end{abstract}

\section{Introduction}
\begin{figure}[t]
\centering
\includegraphics[width=\linewidth]{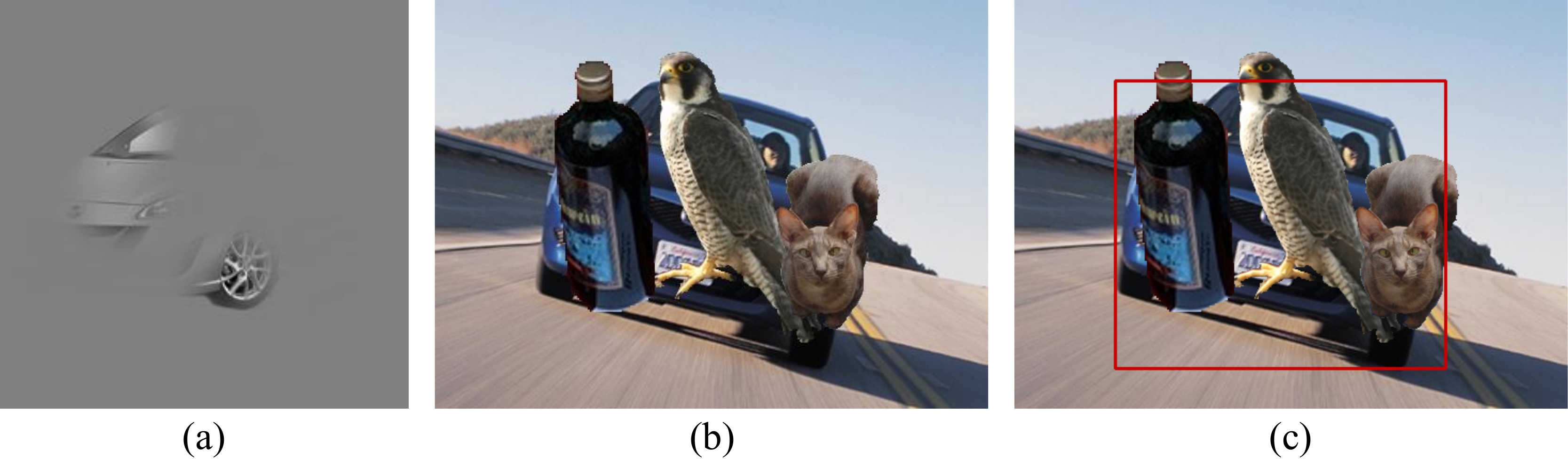}
\caption{Examples of different types of occlusion (on car images). (a) An image of constant mask occlusion used in \cite{tang2018recurrent}. (b) An image of extreme occlusion used to test computational models in our paper. (c) An image of extreme occlusion used in our behavioral experiments.}
\label{fig:dataset}
\end{figure}

Objects in the visual world are occluded much more than objects in typical visual science experiments. The ability to handle occlusion is essential for survival and everyday activities. For instance, in order to safely drive on the road, one must be able to swiftly detect other vehicles and pedestrians in advance even when they are only partially visible. However, both humans and object recognition models are rarely tested on the level of occlusion we encounter in the real world. Several studies have addressed this important issue by investigating real-world object recognition under occlusion \cite{tang2018recurrent,rajaei2018beyond}. These authors successfully developed models that could handle constant mask occlusion as shown in Figure~\ref{fig:dataset}~(a), and produced results consistent with human performance. In this paper, we take a step further and propose a more challenging task of object recognition under extreme occlusion to test humans and object recognition models.

We designed our task using a public occlusion image dataset from the computer vision community \cite{wang2017detecting}. In the proposed task, target objects are heavily occluded by several superimposed irrelevant real-world objects (occluders) with an average target occlusion ratio above 0.6, see Figure~\ref{fig:dataset}~(b).
The biggest difference from previous studies is that both targets and occluders are real objects in real backgrounds. This task provides new insights in two ways. First, compared to testing in an occlusion-free domain, it challenges both humans and object recognition models and better tests the robustness of visual recognition. Second, it provides a way to check whether the ability to handle constant mask occlusion can entail robustness to real-world occlusion.

To begin with, we experimentally measured human performance on our task. Figure~\ref{fig:dataset}~(c) provides an example stimulus used in the behavioral experiments. Humans were very good at recognizing such extremely occluded objects and showed great robustness to occlusion. The results also suggest that our task is feasible and that an ideal object recognition model should be able to accomplish it.


Therefore, we subsequently tested several recent object recognition models on our task. The first model tested is the hierarchically feed-forward model, represented by convolutional neural networks (CNNs) \cite{lecun2015deep}. These models mimic the feed-forward process in biological vision. They achieve impressive performance and can explain some human data in several non-challenging visual tasks \cite{dicarlo2007untangling,yamins2014performance}. However, there is still considerable variability in human neural and behavioral data at the individual image level that CNNs cannot explain \cite{rajalingham2018large,schrimpf2018brain}. Our experiments show that CNNs perform very well without occlusion but their performance is not very good under extreme occlusion, suggesting that they lack robustness to occlusion.

The second model is a hybrid model that combines CNNs with models of recurrent computations. In biological vision, recurrent computations are essential for recognition under occlusion \cite{lamme2000distinct,lamme2002masking,tang2018recurrent,rajaei2018beyond,wyatte2012limits}. \citeA{tang2018recurrent} modelled recurrent computations as lateral connections realized by a Hopfield network, which acted as a content addressable memory \cite{hopfield1982neural}. They used Hopfield networks to store CNN activations of occlusion-free objects and later recover activations of occluded objects. This model improved CNN performance for recognition under constant mask occlusion, but we did not observe improvements under extreme occlusion, implying that the ability to handle constant mask occlusion does not entail robustness to real-world occlusion. 

As a comparison to the models above, we propose a third model designed with the principle of composition. Throughout this paper, we use the term ``composition'' in its traditional sense meaning the process where smaller parts are composed together to form larger parts. This principle is not inherently addressed by CNNs \cite{stone2017teaching}, but it is supported by biological evidence showing that populations of neurons in macaque V4 and IT areas represent complete shapes with aggregates of shape fragments \cite{brincat2006dynamic,pasupathy2002population,yamane2008neural}. Our model uses two stages to recognize objects from parts/subparts in a compositional manner. In the first stage, we consider spatial relations among subparts and detect parts. In the second stage, we consider spatial relations among detected parts and compose them into objects. Both stages are designed to be robust to occlusion. This two-stage model can handle missing parts under occlusion as long as the visible parts conform to reasonable spatial constraints. Our model performed better than the other two models under extreme occlusion with similar error patterns to humans, demonstrating a way to build robustness to occlusion by exploiting object compositional structures.


\section{Task and Dataset}
\subsection{Recognition Under Extreme Occlusion Task}
Object recognition under occlusion involves recognizing targets  occluded by other entities (called occluders). Depending on task specifics, targets could be as simple as letters, digits, and symbols, or as complex as objects in real scenes. Occluders can also vary substantially. There are simple occluders like constant masks and also complex ones like real objects. 

Simple occlusion by constant masks and complex occlusion by real objects are actually treated differently during recognition. For instance, in CNNs, neurons tuned to fur textures may fire on the presence of cats as occluders and distort the CNN activation of the target. However, fewer misguiding neurons are likely to fire on constant masks without textures.  Thus, real objects as occluders are more likely to distort target object recognition by providing irrelevant context.

To test the robustness of visual recognition to real-world occlusion, we propose a difficult task of object recognition under extreme occlusion. Specifically, it involves recognizing vehicles occluded by other irrelevant real occluders, including animals, furniture and other objects, in real backgrounds (see Figure~\ref{fig:dataset}~(b)). This task is challenging because occluders are irrelevant real objects and the occlusion ratio is high, which discourages the use of context during recognition.

\subsection{Training and Testing Dataset}
For the purpose of training computational models to perform our task, we propose that only occlusion-free images should be used. A single object can be occluded in an exponential number of ways and it is unlikely for a limited training set to cover all occluder appearances, positions and so on. Therefore, we used 4049 occlusion-free training images covering five types of vehicles, including \textit{aeroplane}, \textit{bicycle}, \textit{bus}, \textit{car} and \textit{motorbike} from VehicleSemanticPart dataset \cite{wang2017detecting}. 113 different types of object parts, such as car wheels, bicycle pedals and jet engines, are annotated with part identities and bounding box positions.


For testing purposes, we built an occlusion testing set using another 500 images from VehicleOcclusion dataset \cite{wang2017detecting}. To our knowledge, this is the only public occlusion dataset with accurate occlusion annotations of parts and objects. In each image, 2-4 randomly-positioned real occluders are placed onto the single target object (see Figure~\ref{fig:dataset}~(b)). The target occlusion ratio is constrained. 77\% of the images have an occlusion ratio of 0.6-0.8; 18.4\% of the images have an occlusion ratio of 0.4-0.6; 4.6\% of the images have an occlusion ratio of 0.2-0.4. Furthermore, we also created an occlusion-free testing set by collecting the corresponding 500 clean images before occluders were placed. Neither these clean images nor superimposed occluders are met in the training set.

For evaluation metrics, we evaluated human and model performance both quantitatively by recognition accuracy and qualitatively by confusion matrices and representational dissimilarity matrices \cite{kriegeskorte2008representational}.

\begin{figure*}[t]
\centering
\includegraphics[width=\linewidth]{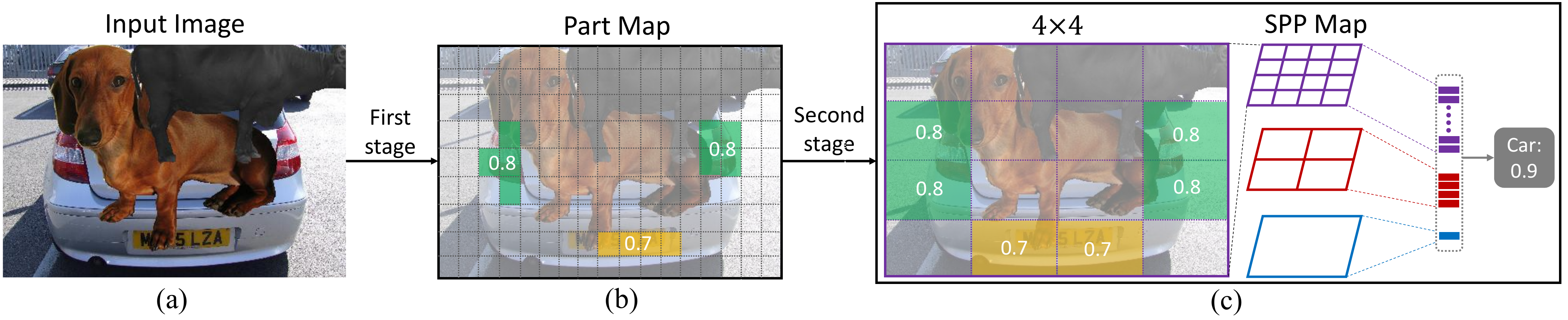}
\caption{The overall framework of our two-stage model. The input image~(a) contains a car whose back trunk is occluded. In the first stage, we detect several parts, including tail lights (green) and the license plate (yellow), and output part maps~(b). In the second stage, we apply spatial pyramid pooling (SPP) to part maps and obtain SPP maps~(c) ($4\times 4$ scale shown here). We consider spatial constraints over parts and aggregate part confidence scores at different scales to determine the object identity.}
\label{fig:model}
\end{figure*}

\section{Behavior Experiments}
\subsection{Participants}
We designed a survey on Amazon Mechanical Turk to collect human responses for the task of object recognition under extreme occlusion. 25 human subjects completed our survey.
\subsection{Procedure}
We had 1,250 human intelligence tasks (HITs) with 20 stimuli in each, so that there were 50 repetitions of the occlusion testing set (500 stimuli). In each HIT, subjects were given unlimited time to observe and respond to 20 stimuli one at a time. The stimuli had red bounding boxes around the targets (see Figure~\ref{fig:dataset}~(c)). Subjects were asked to type the names of the objects in the bounding boxes without knowing that they should belong to the aforementioned five categories. 

\subsection{Data Processing and Exclusions}
We collected 25,000 typed strings as subject responses. There were 785 different strings from the responses and we manually assigned them to the five vehicle categories. We first excluded 300 responses whose corresponding images were oversized for computational models. We further excluded 5,359 responses assigned to either none or more than one category. The rest of 19,341 responses with valid reported and ground-truth category labels were used for data analysis.

\section{Computational Models}


\subsection{CNNs: AlexNet, ResNet and VGG16}
Recently, CNNs achieved impressive performance in object recognition. They use stacked convolutional layers and pooling layers to extract image features for classification. We used AlexNet \cite{krizhevsky2012imagenet}, ResNet \cite{he2016deep} and VGG16 \cite{simonyan2014very} as three representative CNN-based object recognition models and modified them to get the best possible performance. In AlexNet, we substituted fully-connected layers fc6 and fc7 with two equivalent convolutional layers with kernel size $6\times 6$ and $1\times 1$, followed by a global average pooling layer. In VGG16, we also made similar modifications to the fully-connected layers. In ResNet, we used ResNet-18 and substituted the last average pooling layer with a global average pooling layer. These modifications enable the networks to handle inputs of variable sizes. The classification layers were also changed to fit our five object categories. 

\subsubsection{Training and Testing} 
The training and testing images were resized so that the short object edge had 224 pixels. During the training phase, we froze weights in earlier layers and fine-tuned ImageNet pre-trained models. In AlexNet, we froze weights in the first two convolutional layers. In ResNet, we froze weights in the first two residual stages. In VGG16, we froze weights in the first two convolutional stages. The training inputs were randomly cropped image patches of size $224\times 224$ containing at least part of the target objects. During the testing phase, inputs were full-sized images. 

\subsection{Hybrid Model: AlexNet+Hopfield Network}
\citeA{tang2018recurrent} proposed a hybrid model of CNNs with Hopfield networks that improved object recognition under constant mask occlusion and produced results consistent with human performance. Concretely, they adopted a fully-connected Hopfield network with binary threshold nodes. It can store patterns as local minima and later recover incomplete patterns by iteratively processing them until convergence. The capacity of a Hopfield network with N nodes is only about 0.15N memories. When the number of patterns to store increases, local minima are more likely to be spurious minima. We followed the experimental settings from \citeA{tang2018recurrent} and used 4096-dimensional features from the fc7 layer in ImageNet-trained AlexNet as patterns to store and recover in the Hopfield network.

\subsubsection{Training and Testing} During the training phase, we fed a random subset of 500 resized images ($224\times 224$) to ImageNet-trained AlexNet. Following \citeA{tang2018recurrent}, we extracted fc7 features, binarized them with a threshold of 0 and used them to train a Hopfield network with 4096 nodes (implemented in MATLAB's newhop function) and a linear multiclass Support Vector Machine (SVM). We used a small training set due to the limited capacity of the Hopfield network. During the testing phase, we fed images to AlexNet, binarized fc7 features and used them to initialize the Hopfield network. Each node in the network receives weighted inputs from connected nodes and updates its binary state accordingly and synchronously. Converged outputs (timestep=256) are classified using the SVM.

\subsection{Ours: Two-stage Voting Model}
\subsubsection{Motivation} Object appearance and context can change drastically under occlusion yet spatial constraints over objects and parts are largely preserved. With a few parts missing, an occluded object can be recognized as long as the positions of visible parts make sense, that is, if they conform to reasonable spatial constraints. This motivates us to exploit object compositional structures for object recognition under occlusion. We developed a two-stage object recognition model that could utilize these spatial constraints (Figure~\ref{fig:model}). Specifically, in the first stage, we detected different object parts in the images. In the second stage, we considered spatial constraints over objects and parts and used those detected parts to determine object identities. In both stages, we utilized deep networks to capture these spatial constraints via spatial voting mechanisms. The core idea of spatial voting is to detect larger parts by considering spatial relations of smaller parts and gathering their votes, which are confidence scores for their presence. We now elaborate our two stages in more details.

\subsubsection{Stage 1: Part Detection} In the first stage, we want to robustly detect object parts. \citeA{zhang2018deepvoting} developed a robust voting model to detect semantic parts under occlusion. We adopted their method and produced part maps as shown in Figure~\ref{fig:model}~(b), which contained confidence scores for the presence of different object parts at different locations. Following \citeA{zhang2018deepvoting}, we first obtained subparts by clustering corresponding CNN intermediate activations at the \textit{pool-4} layer from a VGG network \cite{simonyan2014very}. We subsequently implemented the spatial voting mechanism as a convolutional layer with kernel size $15\times 15$ on top of the subparts to capture spatial constraints over subparts and parts in a larger spatial region. The intuition for this voting stage is that a partially occluded object part could be robustly detected as long as it gathered enough votes from a set of subparts which conformed to certain spatial constraints in a spatial region.

\subsubsection{Stage 2: Object Recognition} 
\begin{figure}[t]
\centering
\includegraphics[width=\linewidth]{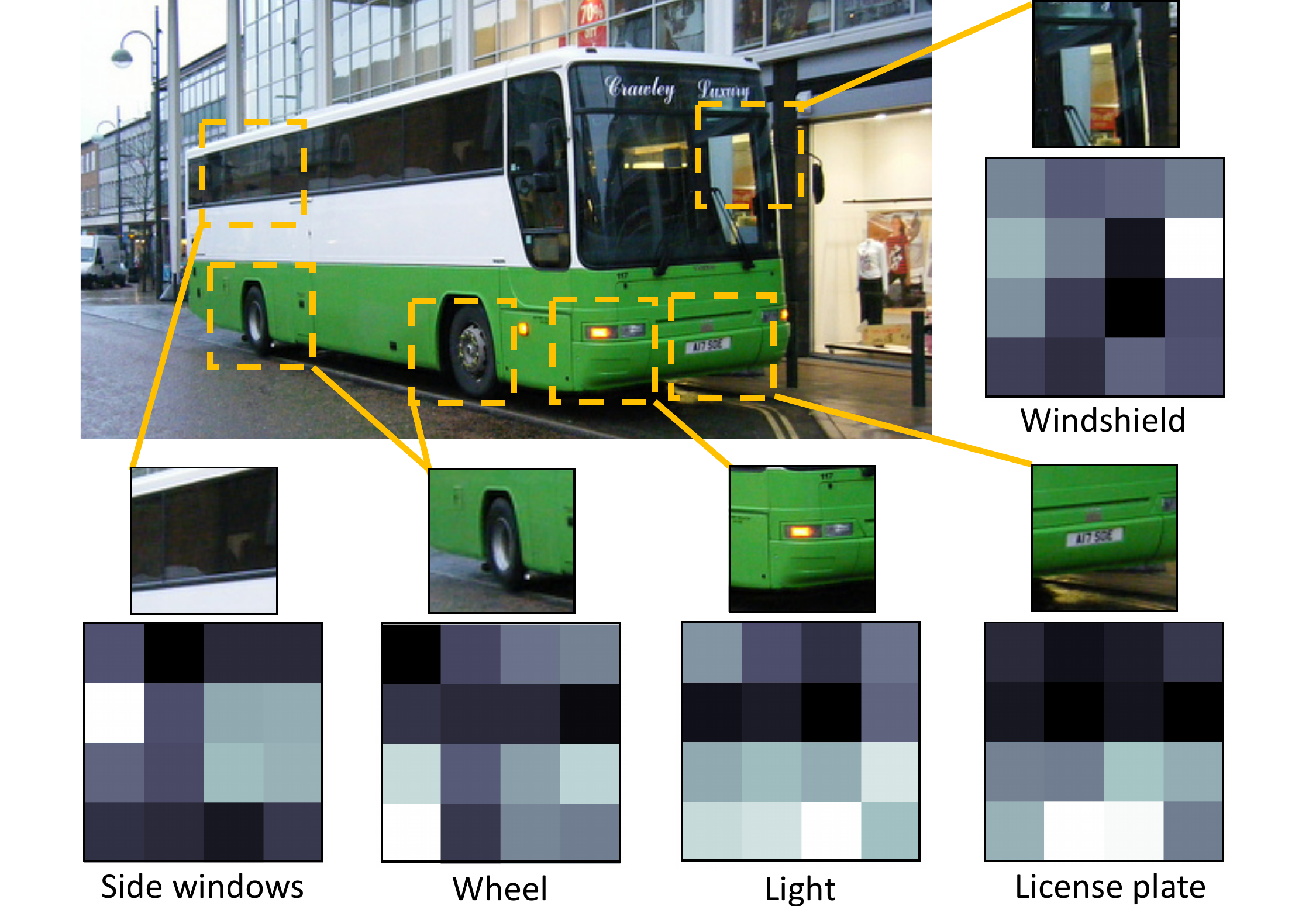}
\caption{Visualization of object-part spatial heatmaps at the $4\times4$ scale for the bus category. Each heatmap shows the learned spatial constraints between the bus and the part. Brighter regions indicate higher voting weights. For instance, a license plate in the lower region of a image often casts a highly weighted vote in favor of the presence of a bus.}
\label{fig:heatmap}
\end{figure}

In this stage, we used part maps as inputs and designed another voting method to aggregate parts to form objects. There are two challenges for our voting method. First, it needs to capture spatial constraints over parts and produce fixed-length vectors for classification. Second, it needs to tolerate within-category variation so that the learned spatial constraints are generalized for most objects in a category. To this end, we applied spatial pyramid pooling (SPP) \cite{lazebnik2006beyond} at three different scales ($4\times 4$, $2\times 2$, $1\times 1$) to the normalized part maps and obtained three SPP maps as shown in Figure~\ref{fig:model}~(c). This method maintained some spatial information and was different from the sliding window pooling of deep networks. Concretely, we evenly divided part maps into $n\times n$ local spatial bins ($n = 4, 2, 1$) and applied max pooling to each bin. SPP maps contained maximum confidence scores for the presence of parts in each spatial bin. Later, we concatenated SPP maps and appended a dropout layer (dropout ratio 0.1) and a fully-connected layer. The dropout layer randomly dropped a subset of part votes during training and improved the robustness of our model under occlusion. The fully-connected layer aggregated part votes and learned spatial constraints over parts at different scales to determine the object identities. The learned weights of this layer are referred to as object-part spatial heatmaps (visualized in Figure~\ref{fig:heatmap}). Both stages in our model were robust to occlusion due to the use of spatial voting and dropout mechanisms, which allowed larger parts to be detected using only a subset of smaller parts under certain spatial constraints.

\subsubsection{Training and Testing}
Training and testing images were resized so that the short object edge had 224 pixels. Two stages were trained separately. When training the first stage, inputs were randomly cropped image patches of size $224\times 224$ containing at least part of target objects. More details are available in \cite{zhang2018deepvoting}. When training the second stage, inputs were parts maps obtained by feeding training images to the first stage. During testing, inputs were full-sized images.

\section{Results and Discussions}
\subsection{Testing without Occlusion}
First, we test three computational models on the task of recognizing occlusion-free vehicles. Human subjects are not tested for this task because it is very easy for humans when they are given unlimited time to recognize vehicles without occlusion.

Both CNNs and our model perform reasonably well on this task (Table~1). Our model has a comparable accuracy (92.9\%), showing that spatial constraints over parts/subparts are useful information for the recognition of occlusion-free objects. However, these results tell little about the robustness of different models given their similar performances. 
\begin{figure*}[t]
\centering
\includegraphics[width=\linewidth]{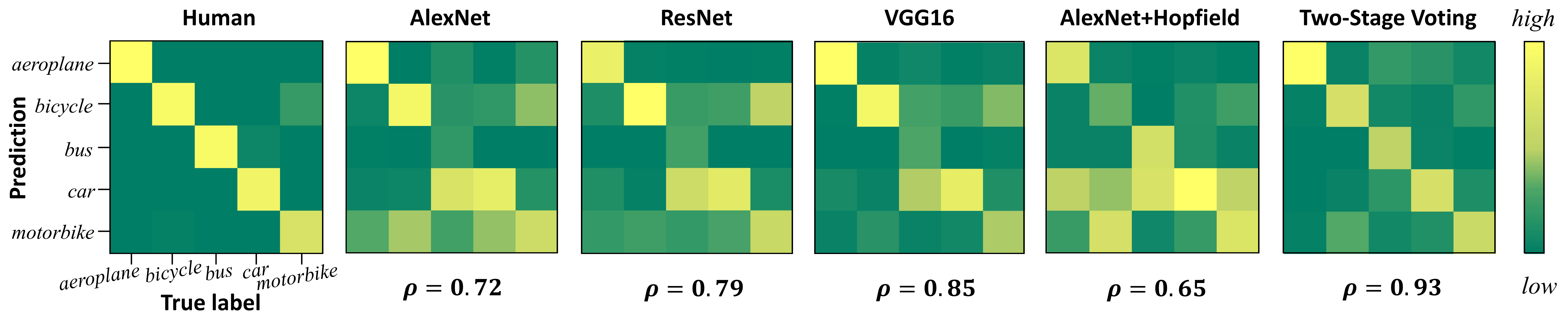}
\caption{Category-level confusion matrices under extreme occlusion. The Pearson correlation coefficients between the human confusion matrix and each model confusion matrix are listed below the matrices.}
\label{fig:confusion}
\end{figure*}

The hybrid model of AlexNet and Hopfield networks gets a relatively lower accuracy without occlusion. When we use SVM to directly classify binarized fc7 features without using the Hopfield network, the accuracy increases from 77.7\% to 85.4\%. This implies that the relatively lower accuracy may be caused by the Hopfield network. In order to check whether the Hopfield network setting was implemented correctly, we followed \citeA{tang2018recurrent} and tested the hybrid model on mask occlusion images from five categories with an average occlusion ratio above 0.7; see Figure~\ref{fig:dataset}~(a) for an example testing image. The use of the Hopfield network increased the accuracy from 40.9\% to 46.8\%, which was qualitatively similar to the improvement reported in \citeA{tang2018recurrent}. This shows that the Hopfield network was implemented correctly and improved object recognition performance under constant mask occlusion. We will further discuss the possible causes of the relatively low accuracy of the hybrid model on our task later with testing results under extreme occlusion.

\begin{table}[t]
\begin{center} 
\caption{Testing Accuracy under No/Extreme Occlusion.} 
\label{sample-table} 
\vskip 0.12in
\begin{tabular}{c|c|c} 
\hline
Humans/Models & w/o occlusion & w/ occlusion\\
\hline
Humans        &  - & \textbf{93.3\%} \\
\hline
AlexNet   &   89.8\% & 50.0\% \\
ResNet           &   90.1\% & 54.0\% \\
VGG16 & \textbf{94.7\%} & 62.6\%\\
AlexNet+Hopfield         &   77.7\% & 46.0\% \\
Two-stage~Voting~(Ours)            &   92.9\% & \textbf{67.0\%} \\
\hline
Ablation 1           &   91.2\% & 47.5\% \\
Ablation 2           &   89.9\% & 58.9\% \\
\hline

\end{tabular} 
\end{center} 
\label{table:acc}
\end{table}

\begin{figure}[t]
\centering
\includegraphics[width=\linewidth]{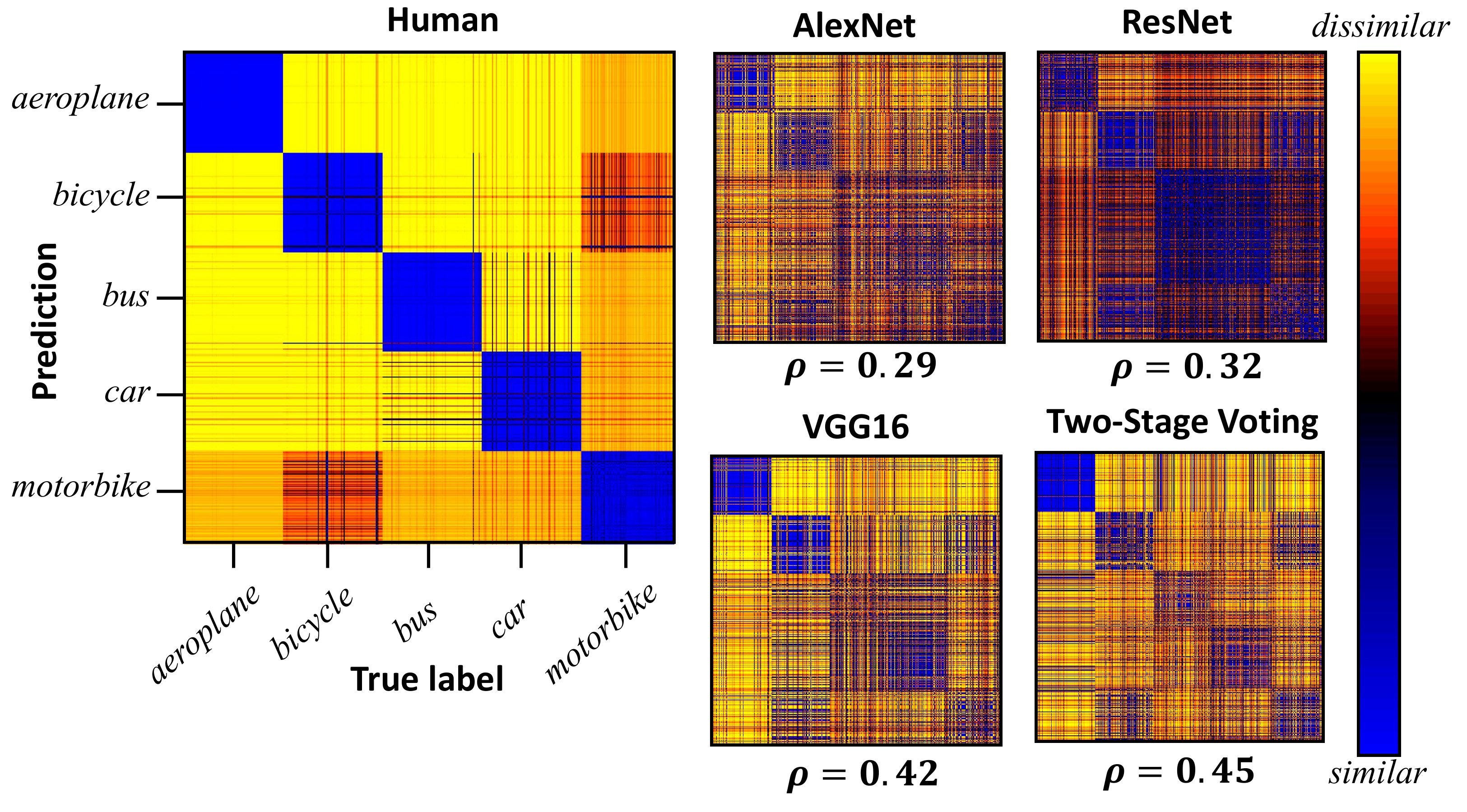}
\caption{Image-level representational dissimilarity matrices (RDMs) under extreme occlusion. Each testing image is characterized by a 5-dimensional categorical distribution obtained from either human response frequencies for five vehicle categories on that image or the Softmax output in the final layer of each model. The dissimilarity between two images is measured as the Euclidean distance between two vectors representing the associated categorical distributions. The Pearson correlation coefficients between the human RDM and each model RDM are listed below the matrices.}
\label{fig:img-confusion}
\end{figure}

\subsection{Testing under Extreme Occlusion}
We further test humans and these models on recognizing objects under extreme occlusion with our occlusion testing set. 

Table~1 shows that humans have very high accuracy at recognizing occluded vehicles and are robust to extreme occlusion. It also confirms that our task of object recognition under extreme occlusion is feasible and the information in these occlusion images is sufficient to determine object identities.


For CNNs, the accuracy is relatively low (Table~1). Despite their good performance without occlusion, CNNs do not manifest robustness under extreme occlusion as humans do. Our results support previous findings that CNN activation is not inherently compositional and cannot explicitly address contextual and non-contextual information \cite{stone2017teaching}.

\begin{figure*}[t]
\centering
\includegraphics[width=\linewidth]{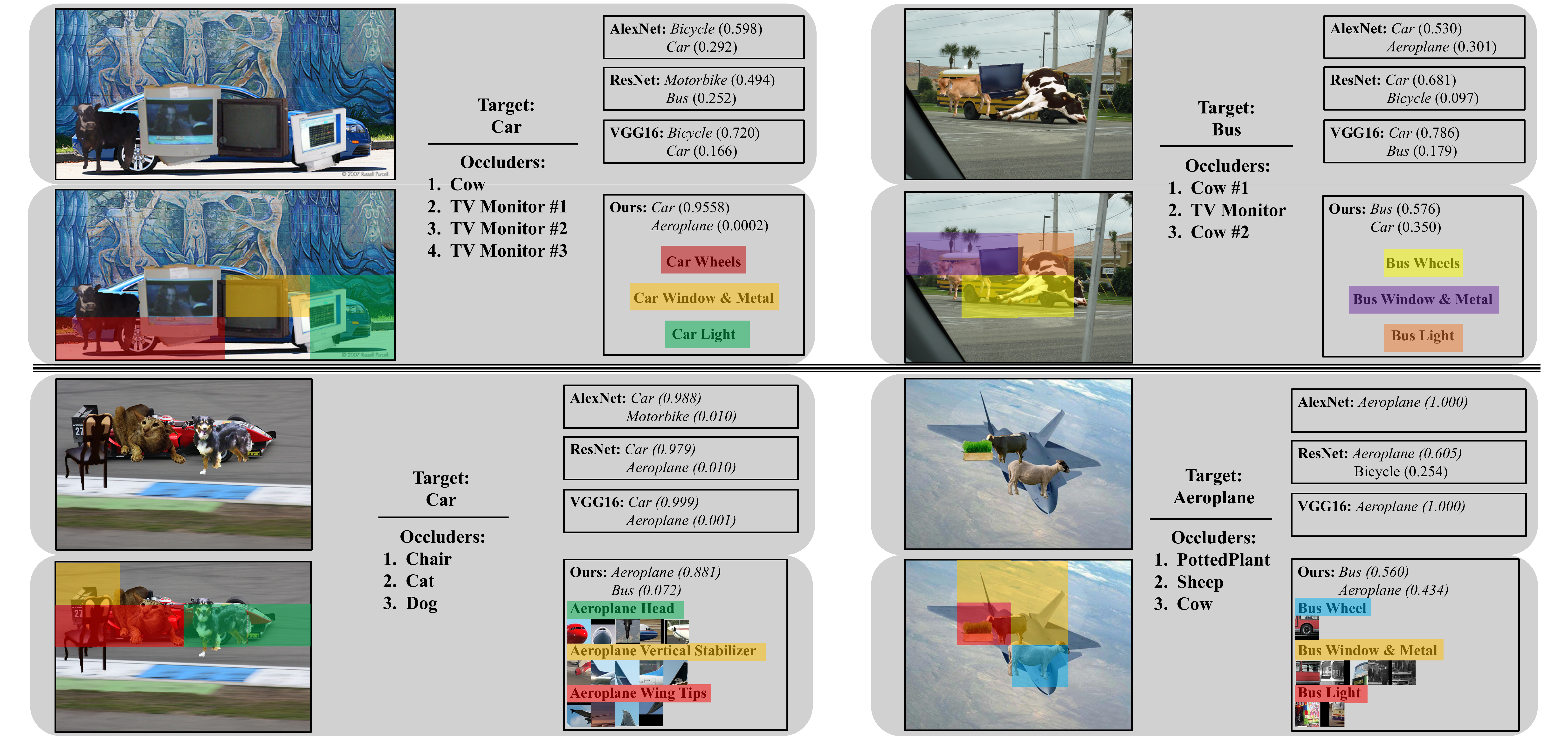}
\caption{Representative improvements (top two) and errors (bottom two) from our model. In the top two cases, our model produced correct object labels by exploiting spatial relations among detected object parts. In the bottom left case, our model misclassified the race car as an aeroplane partly because they are similar in the aerodynamic design. In the bottom right case, our model incorrectly detected bus parts and was subsequently misled by these false part detection results.}
\label{fig:examples}
\end{figure*}


For the hybrid model, we do not observe performance gains compared to CNNs under extreme occlusion. If we use SVM to classify binarized fc7 features without using Hopfield networks, the accuracy rises from 46.0\% to 48.6\%. This result together with previous ones shows that the Hopfield network did not improve performance on our task, either without occlusion or under extreme occlusion, although it improved performance under constant mask occlusion. There are two possible reasons. First, it may require more representative training features because our dataset is more complex than the one from \citeA{tang2018recurrent}. However, given the limited capacity of Hopfield networks, too many training features may result in more spurious minima. Second, mask occlusion may only suppress some neurons in CNNs while real occluders can activate additional misguiding neurons, making it hard for pattern recovery. Thus, the ability to handle constant mask occlusion does not entail robustness to real-world occlusion.

Our model outperforms other models under extreme occlusion in terms of accuracy. We further compare the performance of humans and different models by analyzing their category-level confusion matrices (Figure~\ref{fig:confusion}) and image-level representational dissimilarity matrices (RDMs) (Figure~\ref{fig:img-confusion}). These provide a better way to qualitatively compare the robustness of human vision and these computational models.  Although the accuracy of our model is still lower than humans, it shows greater robustness than other models and produces the most similar results to humans. Figure~\ref{fig:examples} also shows some representative improvements and errors from our model. The performance of our model suggests that spatial constraints over parts are important cues for object recognition under occlusion, and that the principle of composition is a promising solution for bridging the gap between the robustness of human vision and these computational models.

\subsection{Ablation Experiments}
Finally, we show the effectiveness of each voting stage in our model. We substituted each stage respectively with alternative models to do the same task with the same supervision.

In the first experiment (Ablation 1 in Table 1), we substituted the first voting stage with Faster-RCNN \cite{ren2015faster}, a state-of-the-art object detection model. We trained Faster-RCNN to detect parts and obtained SPP maps. Later, we trained our second stage and tested the whole model. The accuracy changed little without occlusion but dropped from 67.0\% to 47.5\% under extreme occlusion. As \citeA{zhang2018deepvoting} pointed out, it is difficult for proposal-based detection methods including Faster-RCNN to extract good proposals under occlusion and even with correct proposals, the classifier may still go wrong due to the presence of occluders. This result further confirmed the robustness of the first stage model on detecting parts under occlusion.

In the second experiment (Ablation 2 in Table 1), we substituted the second stage with a bag-of-words module where all spatial relations were discarded by a global max pooling layer. We concatenated the highest confidence score in each part map into a vector and appended a dropout layer and a fully-connected layer. We trained the bag-of-words module and tested the whole model. The accuracy changed little without occlusion but dropped from 67.0\% to 58.9\% under extreme occlusion, implying the effectiveness of spatial voting under occlusion. Figure~\ref{fig:heatmap} also shows that the learned spatial constraints are meaningful. Wheels and windshields often appear in lower and higher regions of bus images respectively.

Together, the results suggest that the higher accuracy of our model is not purely a result of additional part-level supervision but also due to the use of object compositional structures.

\section{Conclusion}

Occlusion is often present in everyday visual tasks yet humans and models are rarely tested under real-world occlusion. We proposed a task of object recognition under extreme occlusion and tested humans and models, including CNNs, a hybrid model of CNNs with Hopfield networks and our two-stage voting model. Our findings lead us to three conclusions.

First, testing under extreme occlusion can better reveal the robustness of visual recognition than testing without occlusion. Object recognition models that can compete with humans in the occlusion-free domain may not show the same robustness under extreme occlusion as humans do.

Second, the ability to handle constant mask occlusion does not entail robustness to real-world occlusion. Different types of occlusion may alter context differently yet object inherent structures could still be exploited for recognition purposes.

Third, the performance of our model is better and more correlated with human results under occlusion, suggesting that the principle of composition is a possible solution for building robustness to occlusion as demonstrated by human vision.


\section{Acknowledgments}
We thank Tal Linzen, Dan Kersten, Tom McCoy and the JHU CCVL group for helpful comments.
This work was supported by ONR with grant N00014-19-S-B001.

\bibliographystyle{apacite}

\setlength{\bibleftmargin}{.125in}
\setlength{\bibindent}{-\bibleftmargin}

\bibliography{CogSci_Template}

\end{document}